%% file: emnlp2023.tex
\pdfoutput=1

\documentclass[11pt]{article}
\usepackage{authblk}
\usepackage{EMNLP2023}

\usepackage{times}
\usepackage{latexsym}

\usepackage[T1]{fontenc}

\usepackage[utf8]{inputenc}
\usepackage{mathrsfs}
\usepackage{mathtools}
\usepackage{microtype}
\usepackage{amsmath,amssymb,amsfonts}
\usepackage{algorithm}
\usepackage{algorithmicx}
\usepackage{algpseudocode}
\usepackage{multirow}
\usepackage{graphicx}
\usepackage{booktabs}
\usepackage{makecell}
\usepackage{colortbl}
\usepackage{xcolor}
\usepackage{latexsym}
\usepackage{microtype}

\usepackage[misc]{ifsym}

\definecolor{good_color}{rgb}{0.88, 1, 0.88}
\definecolor{bad_color}{rgb}{1, 0.88, 0.88}
\usepackage{inconsolata}

\newcommand{\stacklines}[3][0.5ex]{\begin{tabular}[t]{@{}c@{}}#2\\[#1]#3\end{tabular}}

\title{Let GPT be a Math Tutor: Teaching Math Word Problem Solvers with Customized Exercise Generation}

\author[1]{\textbf{Zhenwen Liang} \textsuperscript{\tiny \Letter}}
\author[1]{\textbf{Wenhao Yu}}
\author[2]{\textbf{Tanmay Rajpurohit}}
\author[3]{\\ \textbf{Peter Clark}}
\author[1]{\textbf{Xiangliang Zhang}}
\author[3]{\textbf{Ashwin Kaylan} \textsuperscript{\tiny \Letter}} 
\affil[1]{University of Notre Dame, \texttt{zliang6@nd.edu}}
\affil[2]{Georgia Institute of Technology}
\affil[3]{Allen Institute for AI, \texttt{\;ashwinkv@allenai.org}}

\begin{document}
\maketitle
\begin{abstract}

In this paper, we present a novel approach for distilling math word problem solving capabilities from large language models (LLMs) into smaller, more efficient student models. Our approach is designed to  consider the student model's weaknesses and foster a tailored learning experience by generating targeted exercises aligned with educational science principles, such as knowledge tracing and personalized learning. 
\textcolor{black}{Concretely, we let GPT-3 be a math tutor and run two steps iteratively: 1)  assessing  the student model's current learning status on a GPT-generated exercise book, and 2) improving the student model by  training it with    tailored exercise samples  generated by GPT-3. }
Experimental results reveal that our approach outperforms LLMs \textcolor{black}{(e.g., GPT-3 and PaLM)} in accuracy across three distinct benchmarks while employing significantly fewer parameters. Furthermore, we provide a comprehensive analysis of the various components within our methodology to substantiate their efficacy.
\end{abstract}

\section{Introduction}

Math word problems (MWPs) are an essential aspect of mathematical education and a critical skill to develop for individuals across various domains in life \cite{hegarty1995comprehension}. MWP solving requires the ability to comprehend natural language, extract relevant information, and perform mathematical reasoning to solve the given problem. Research in MWP solvers has gained considerable interest due to the ubiquity of such problems in daily life, ranging from finance and scheduling to engineering and science \cite{koedinger2004real}. Developing AI agents capable of solving MWPs demonstrates an advanced understanding of the interplay between natural language processing and mathematical reasoning, paving the way for practical applications in education, decision-making, and more \cite{mukherjee2008review}. The format of a typical MWP involves a textual description of a problem scenario, which needs to be translated into a mathematical expression (typically an equation) that can be solved to obtain the answer. MWP solving was presented as a task for artificial intelligence several decades ago \cite{fletcher1985understanding}. Previous fine-tuned methods usually apply Seq2Seq models \cite{wang2017deep,wang2018translating,xie2019goal,zhang2020graph,liang2022mwp}. In recent years, large language models (LLMs) such as GPT-3 \cite{brown2020language} and PaLM \cite{chowdhery2022palm} exhibit strong reasoning ability with the help of chain-of-thought (CoT) prompting \cite{wei2022chain,wang2022self,chen2022program}, which achieves striking performance on MWP solving and outperforms fine-tuned state-of-the-art (SOTA) solvers by a large margin. 
\\ \\

\begin{figure}
\centering 
\includegraphics[width=0.48\textwidth]{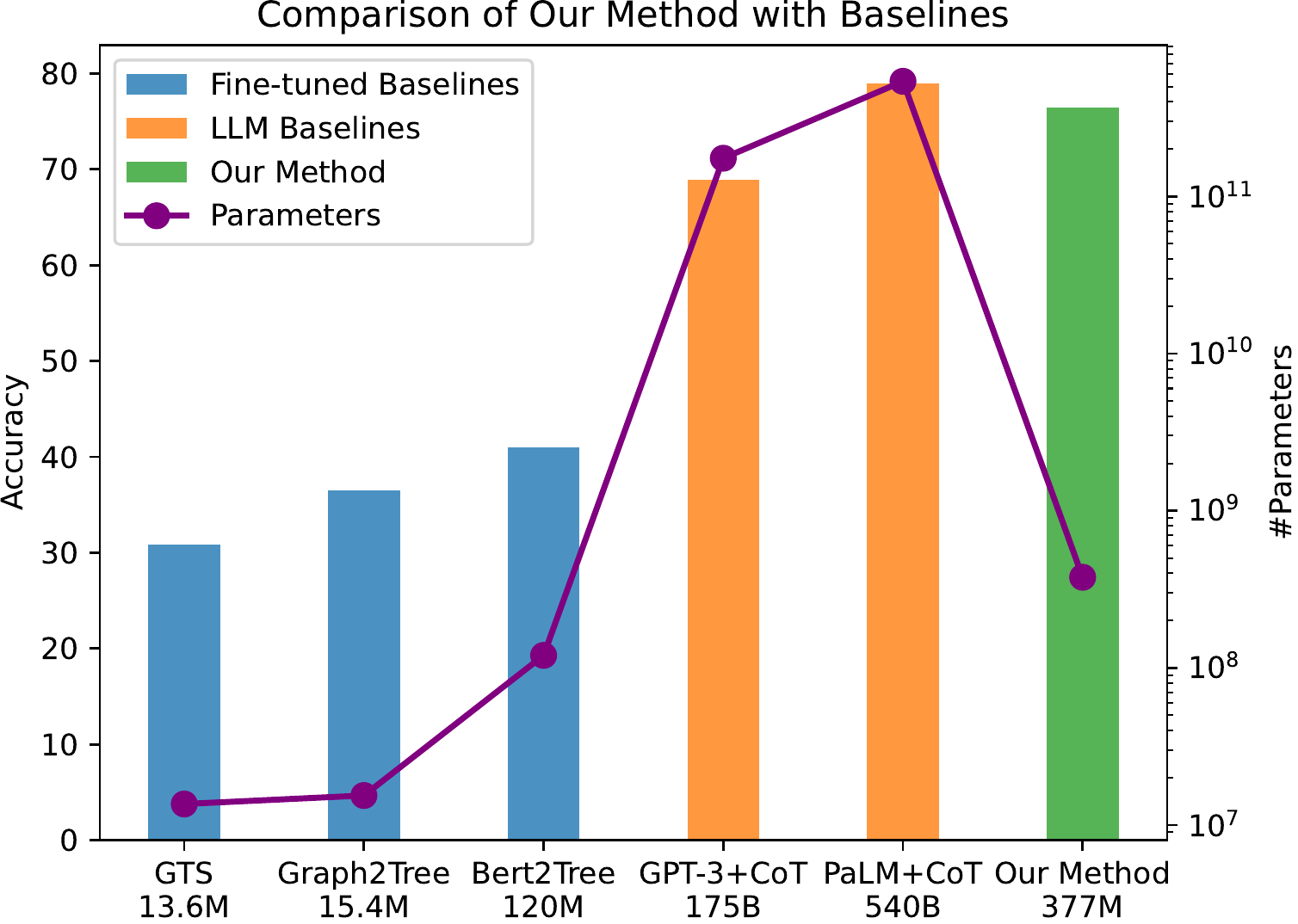} 
\caption{Accuracies vs model sizes for representative baselines and our approach on SVAMP dataset. Our method achieves competitive performance with LLMs with significantly fewer parameters.} 
\label{fig:intro} 
\end{figure}

\vspace{-0.8cm}
In recent years, large language models (LLMs) have made remarkable strides in solving MWPs. However, the substantial number of parameters in LLMs results in computational inefficiency, and necessitates API availability for reproducibility. To mitigate these issues, a natural solution is distilling the knowledge from LLMs into smaller, more efficient student models. The majority of prior research has emphasized using the "explanation" component of the CoT approach as the distilled knowledge \cite{ho2022large, li2022explanations, shridhar2022distilling, magister2022teaching}. Nonetheless, these methodologies exhibit certain limitations. Firstly, small student models may struggle to understand CoT explanations, potentially impeding their learning efficacy. Evidence from \cite{wei2022chain, weiemergent} indicates that only exceptionally large language models possess the capability to perform CoT during reasoning. As a result, many student models \cite{ho2022large, li2022explanations, magister2022teaching} trained on explanations do not attain satisfactory accuracy in MWP-solving tasks. Secondly, those distillation processes lack feedback from the student model to the LLM teacher, and thus, neglect to consider the knowledge state of the student.

In this paper, we also concentrate on harnessing the strengths of LLMs to coach smaller, more efficient MWP solvers, but introduce a novel method for addressing the limitations of previous approaches: \textbf{C}ustomized \textbf{E}xercise for \textbf{MA}th \textbf{L}earning (CEMAL). We reframe the problem by shifting our focus from providing explanations for existing exercises (i.e., training set) to \textit{identifying the student model's learning needs} and \textit{generating new exercises tailored to them.} This approach offers several advantages: (1) it does not impose CoT ability requirements on small models, allowing them to learn more effectively, (2) it takes into account the learning status of the student model during training.

In fact, our approach CEMAL seamlessly aligns with two fundamental tasks in educational science: \textit{knowledge tracing} and \textit{personalized learning}. Knowledge tracing pertains to monitoring and modeling a student's evolving knowledge state over time \cite{corbett1994knowledge,abdelrahman2023knowledge}. This process enables the identification of a learner's strengths and weaknesses, usually by exercises, thereby facilitating the generation of tailored educational experiences. Personalized learning is also of vital importance \cite{hattie2007power,grant2014personalized}. Being cognizant of the student model's learning status ensures the optimally designed exercises are generated to address the model's specific weaknesses. By monitoring the learning progress, our proposed method can dynamically adapt to the student model's evolving knowledge state, fostering more effective learning outcomes. In our method, we integrate knowledge tracing and learning status into the distillation process to establish a robust connection between the LLM teacher and the student model, yielding a more interactive and customized learning experience. Consequently, this approach substantially enhances the student model's problem-solving capabilities. As illustrated in Figure \ref{fig:intro}, our knowledge distillation approach achieves competitive accuracy on the SVAMP dataset, but employs significantly fewer parameters compared to state-of-the-art LLMs, such as GPT-3+CoT and PaLM+CoT \cite{wei2022chain}.


Our contribution can be summarized as follows:

\begin{itemize}
\item We propose a novel method named CEMAL that utilizes LLMs to generate additional data in the form of targeted practice problems, addressing the student model's weak areas.
\item Our approach is evaluated on multiple MWP datasets, including both in-distribution (ID) and out-of-distribution (OOD) tests \cite{koncel2016mawps,miao2020diverse,patel2021nlp}. We show that our method is significantly effective in improving student models under the OOD setting.

\item The experimental results demonstrate that our method achieves state-of-the-art accuracy, significantly outperforming fine-tuned baselines. Notably, the student model trained with our method even surpasses LLMs with CoT prompting, despite having significantly fewer parameters.

\end{itemize}

\section{Related Work}

\subsection{Math Word Problem Solving}
After many years of research on rule-based algorithms \cite{hosseini2014learning,mitra2016learning} and semantic parsing methods \cite{shi2015automatically,huang2017learning}, deep learning has become the predominant technique for solving MWPs, thanks to its superior performance and better generalization ability. Deep neural solver (DNS) \cite{wang2017deep} was among the first to apply Seq2Seq models with RNN encoder and decoder for MWP solving, and subsequent research has mainly explored different structures of RNN-to-RNN solvers \cite{wang2018translating,liu2019tree,xie2019goal,li2019modeling,zhang2020graph,liu2020reverse}. More recently, pre-trained language models such as BERT \cite{devlin2019bert} and RoBERTa \cite{liu2019roberta} have demonstrated remarkable language understanding abilities, leading researchers to replace MWP encoders with these models \cite{li2021seeking,huang2021recall,shen2021generate,patel2021nlp,liang2022mwp}, resulting in significant accuracy improvements. Recently, LLMs have shown great success in MWP solving, with much superior accuracy compared to fine-tuned baselines, simply by being provided with a few CoT examples of the problem-solving processes. Interestingly, researchers also found that LLMs can reason with zero-shot prompts such as "Let's think step by step" \cite{kojima2022large}. Currently, numerous studies have been conducted to improve the performance of the chain-of-thought prompting, including recent works by \cite{shi2022language,wang2022self,chen2022program,lu2022dynamic,zhou2022least,diao2023active}.

In this paper, we leverage the great success of LLMs to facilitate the training of the student solver. Our approach utilizes an efficient fine-tuned solver as its backbone and yields even superior performance to LLMs, further pushing the limit of deep learning methods in MWP solving.

\subsection{Large Language Models for Knowledge Distillation and Data Generation}
In recent years, there has been a surge of interest in knowledge distillation from LLMs to smaller models. Due to the unavailability of model structure for LLMs, their application is often limited to prompt design and subsequent data generation. Therefore, data generation has become an integral part of knowledge distillation in the context of LLMs. Several studies have investigated the potential of LLMs in knowledge distillation and data generation. For instance, PromDA \cite{wang2022promda} applies prompt-based data augmentation to low-resource natural language understanding tasks, and AugESC \cite{zheng2022augesc} leverages the GPT-J \cite{gpt-j} model and utilizes publicly available dialog posts to trigger conversations on various topics. Then, \citet{west2022symbolic} employs a generalized LLM to create common sense knowledge graphs, while WANLI \cite{liu2022wanli} combines LLM-generated examples with human evaluation to establish diverse natural language inference datasets. Additionally, ZeroGen \cite{ye2022zerogen} proposes a zero-shot learning approach by generating datasets using pre-trained LLMs, ProGen \cite{ye2022progen} uses the quality of generated samples as feedback to LLMs to improve the data generation, and \citet{shao2023synthetic} feeds chain-of-thought demonstrations to LLMs and targets generating more exemplars for in-context learning.

In the domain of MWP solving, several studies have also been conducted with the objective of distilling the reasoning capability of LLMs into smaller solvers by employing chain-of-thought explanations. \cite{ho2022large} introduces Fine-tune-CoT, which uses LLMs to generate reasoning step instances, subsequently facilitating the fine-tuning of smaller models. \cite{li2022explanations} explores three explanation generation strategies and incorporates them into a multi-task learning framework tailored for compact models. \cite{magister2022teaching} assesses the efficacy of chain-of-thought explanations in the training of a small model across three disparate tasks, namely arithmetic reasoning, commonsense reasoning, and symbolic reasoning. Furthermore, \cite{shridhar2022distilling} presents Decompositional Distillation, an approach that segments the problem into subproblems to enhance smaller models' performance.

In contrast to these existing works, our proposed knowledge distillation approach in MWP solving is unique in that it does not focus on the chain-of-thought explanation and it takes into account the learning status of the student model and generates exercises that tailor to the specific weaknesses of the student. Our approach bridges the gap between knowledge distillation and data augmentation in the context of math word problem solvers, allowing student models to improve their problem-solving capabilities more effectively.

\section{Approach}

\subsection{Problem Definition}
Our objective is to train a student Math Word Problem (MWP) solver with the assistance of large language models (LLMs). The student MWP solver takes a textual description $W$ as input and produces an equation $A$ as output, which indicates the solution process to arrive at the final answer. We represent quantities in $W$ and $A$ using placeholders following number mapping \cite{wang2017deep}, which unifies the representation of quantities in different MWPs. Specifically, $W$ and $A$ do not contain the actual values of quantities, but are denoted as ${N_1, N_2, ..., N_k}$, where $N_i$ refers to the $i$-th number, and $k$ is the maximum number of quantities in $W$ and $A$. This approach offers two advantages: (1) it is a widely used data pre-processing technique in MWP solving that unifies the representation of quantities and reduces the vocabulary size, and (2) during exercise generation, where the goal is to generate MWP variants, number mapping prevents variants generated solely by changing the values of quantities.

\begin{figure}
\centering 
\includegraphics[width=0.48\textwidth]{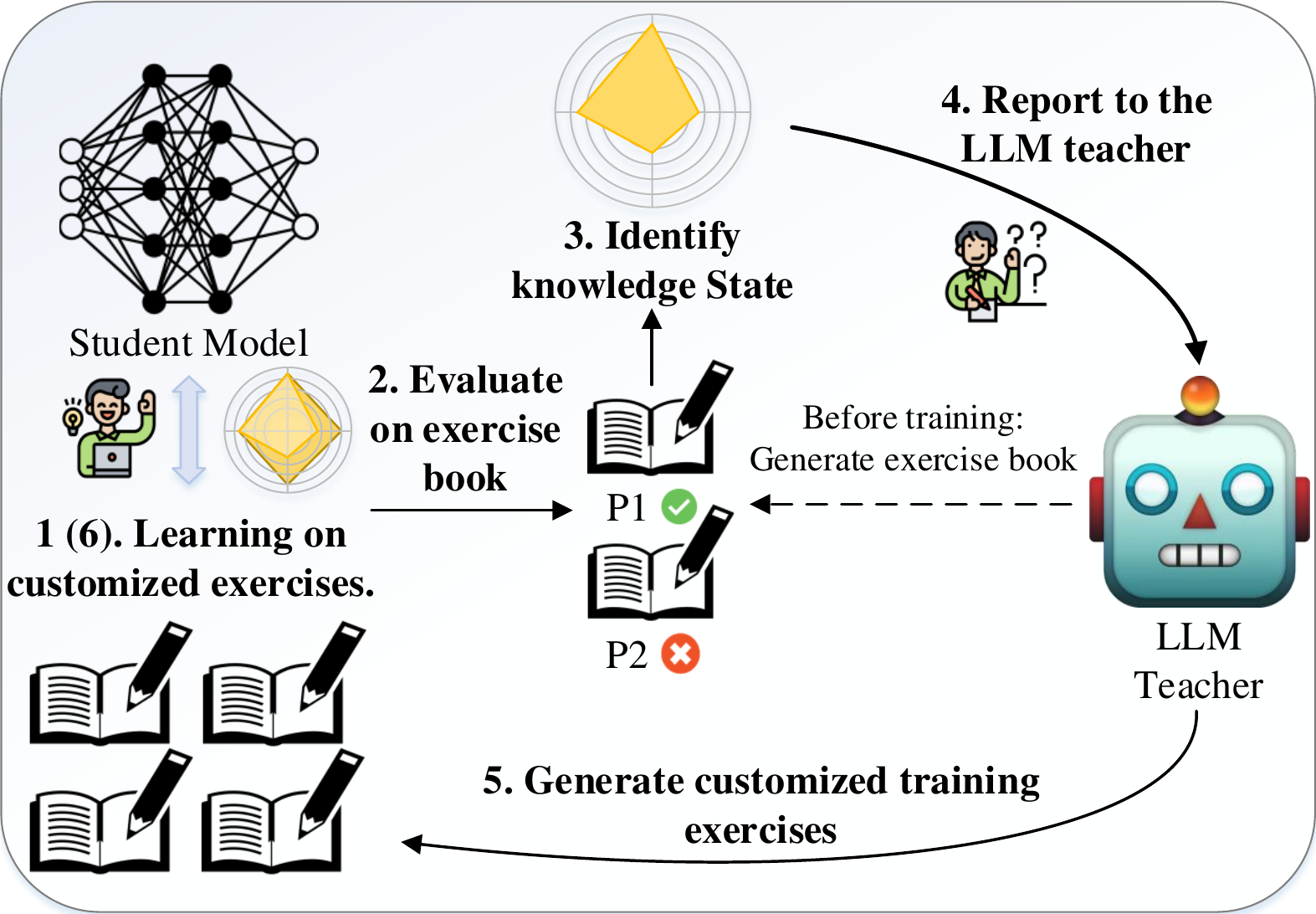} 
\caption{This figure shows the overall iterative framework of CEMAL. After one round of training, the student, which is a small MWP solver, is evaluated by exercises provided by an LLM teacher. Subsequently, LLM generates customized exercises that target the student's knowledge state and weaknesses, thereby facilitating a customized improvement in their overall performance.} 
\label{fig:framework} 
\end{figure}

\subsection{Training Workflow}
\label{notation}
Our proposed training workflow is an iterative and progressive framework that integrates LLMs into the training procedure of student MWP solvers as shown in Figure \ref{fig:framework}. Before training, we augment the training set $\mathcal{T}_{train}$ by generating $m$ and $n$ times more problems than the original training set. One set of the augmented problems is used to enlarge the training set at the beginning to push the limit of the student solvers, while the other set is used to formulate the exercise book $\mathcal{X}$. During training, we perform model validation on $\mathcal{X}$ and select the problems that the solver is unable to solve, which reflect the knowledge state of the student at that time. Subsequently, we generate customized exercises and incorporate them into the original training set $\mathcal{T}_{train}$ to update it. Specifically, we generate $k$ exercises for every source MWP. This iterative process allows for the continuous expansion of the training set and leads to enhanced performance of the student solver. Furthermore, we introduce a threshold $\lambda$ to control the proportion of targeted generation compared to random generation. A detailed description of $\lambda$ can be found in Section \ref{lambda}. The pseudo-code of our approach is located in Alg. \ref{alg:algorithm}. This framework closely mimics the knowledge tracing method in human learning and thus holds promise for enhancing the effectiveness of future educational practices.

\begin{algorithm}[t]
\small
\caption{Training Framework}
\label{alg:algorithm}
\textbf{Input}: Training Set $\mathcal{T}_{train}$, Exercise Generation Function $EG$, Exercise Book $\mathcal{X}$, Student Solver $S$ \\
\textbf{Parameters}: Probability Threshold $\lambda$, number of generations $m,n,k$ \\
\begin{algorithmic}[1] 
\For{Each $W$-$A$ pair in $\mathcal{T}_{train}$} \Comment{Before Training}
\State $\mathcal{T}_{train} \gets \mathcal{T}_{train} + EG(W,A,m)$ \Comment{Initial Augmentation by m times}
\EndFor
\For{Each $W$-$A$ pair in $\mathcal{T}_{train}$}
\State $\mathcal{X} \gets EG(W,A,n)$ \Comment{Generate an exercise book with n times the size of the training set}
\EndFor

\While{Training} \Comment{Training starts}
\For{Each $W$-$A$ pair in $\mathcal{T}_{train}$}
\State Train Solver $S$ with $W$-$A$
\EndFor
\For{Each $W$-$A$ pair in $\mathcal{X}$}
\If{$S$ cannot solve $W$}
\State $p \gets U \sim \text{Uniform}(0,1)$
\If{$p \leq \lambda$}
\State $\mathcal{T}_{train} \gets \mathcal{T}_{train} + EG(W,A,k)$
\Else
\State Randomly sample a $\Tilde{W}$ - $\Tilde{A}$ pair from $\mathcal{X}$
\State $\mathcal{T}_{train} \gets \mathcal{T}_{train} + EG(\Tilde{W},\Tilde{A},k)$
\EndIf
\EndIf
\EndFor
\EndWhile
\end{algorithmic}
\end{algorithm}

\subsection{Backbone of Student Solver}

In this paper, we employ a Seq2Seq model with the Goal-driven Tree-based Solver (GTS) \cite{xie2019goal} as our decoder, which has been widely applied in MWP solving and shown to outperform Transformer decoders \cite{lan2022mwptoolkit}. For the encoder part, we use three different backbones - LSTM \cite{hochreiter1997long}, RoBERTa-base, and RoBERTa-large \cite{liu2019roberta} to demonstrate the backbone-agnostic nature of our approach.

\subsection{Exercise Generation}
\label{method}
Our method generates targeted exercises and their answers based on a given source problem-answer pair, which is similar to data augmentation. We get the inspiration from a previous work on analogical reasoning \cite{liang2022analogical}, which found that two kinds of analogies - problem analogy and solution analogy - can help models better understand MWPs.

For problem analogy, we generate problems with similar problem descriptions but different solutions. Our method perturbs not only questions but also considers context, and we prompt LLMs with few-shot examples to generate MWP variants. In contrast to previous studies that used human annotators \cite{patel2021nlp,yang2022unbiased}, our approach is automated and scalable. For solution analogy, we generate problems with different keywords in the problem description but a similar solution to the source MWP. Interestingly, we find that LLMs can achieve this generation in a zero-shot manner by simply prompting them to generate problems similar to the source problem.

This exercise generation is deployed at two different steps in our proposed training framework. First, we generate an exercise book by augmenting the entire training set, which serves as a validation set to identify the weaknesses of the student solver. We do not directly use the training set to validate the student solver because we want a more diverse validation set to comprehensively evaluate the student model. Additionally, the solver may memorize the MWPs in the training set instead of fully understanding them, as previous research has found that slightly perturbed training MWPs can cause a solver to fail \cite{liang2021solving}. Therefore, validation on the exercise book, which contains many variants of the training set, provides a more robust evaluation.

Secondly, after identifying the problems that the student solver cannot solve from the exercise book, we use them as the source to generate customized exercises and add them to the previous training set. In this way, the size of the training set grows progressively and covers more knowledge, leading to a stronger student MWP solver.

We empirically find that some of the problems generated by LLMs may be of low quality, in the wrong format, or repetitive, and require filtering before being used. We provide a case study including positive and negative examples in Section \ref{case}. In our experiment, we filter out about 30\% of problems with the wrong formats during the generation process. 

\subsection{Targeted Generation vs. Random Generation}
\label{lambda}
Our exercise book is created by augmenting and diversifying the training set, which effectively identifies the weaknesses of the student solver during its learning process. 
An intuitive comparison experiment to our approach is randomly generating exercises for the student solver during learning. To better formulate different generation strategies and evaluate the effectiveness of our method, we define a probability threshold $\lambda$ to control the proportion of targeted generation compared to random generation when generating the augmentations of the training set. A detailed analysis on different strategies can be found in Section \ref{strategy}.

\input{./tbl_main.tex}
\section{Experiments}
\subsection{Datasets}
\paragraph{MAWPS} The MAWPS dataset \cite{koncel2016mawps} is an aggregation of 2,373 English Math Word Problems (MWPs) from various sources, including AddSub \cite{hosseini2014learning}, SingleOp \cite{roy2015reasoning}, MultiArith \cite{roy2015solving}, SingleEq \cite{koncel2015parsing}, and SimulEq \cite{kushman2014learning}. We employ a 5-fold cross-validation for the evaluation on this dataset.

\paragraph{ASDiv-a} ASDiv \cite{miao2020diverse} is an English MWP dataset designed to exhibit a more diverse range of language patterns and problem types, comprising 2,305 MWPs. In accordance with prior studies \cite{patel2021nlp,lan2022mwptoolkit}, we select the arithmetic subset, ASDiv-a, which contains 1,218 MWPs and utilizes a 5-fold cross-validation method for evaluation.

\paragraph{SVAMP} The SVAMP dataset \cite{patel2021nlp} consists of 1,000 English MWPs generated by introducing challenging variations to existing problems. We adopt the two evaluation settings proposed in \cite{patel2021nlp}. The first setting employs a 5-fold cross-validation approach on the 1,000 MWPs, incorporating MAWPS \cite{koncel2016mawps} and ASDiv-a \cite{miao2020diverse} as additional training data for each fold. In the second setting, MAWPS and ASDiv-a serve as the training set, while SVAMP is used as the testing set.

\paragraph{Evaluation Setting} We categorize the evaluations on MAWPS, ASDiv-a, and the first setting on SVAMP as in-distribution (ID) tests, as they all involve 5-fold cross-validation on a specific dataset. Conversely, the second setting on SVAMP is considered an out-of-distribution (OOD) test, given that the training and testing sets originate from distinct sources.

\subsection{Implementation}
In this work, we conducted all experiments using an NVIDIA RTX A6000 graphics card, implemented in Python using the PyTorch framework. The training was performed for 50 epochs with a batch size of 16, using the AdamW \cite{kingma2014adam,loshchilov2018decoupled} optimizer with an initial learning rate of 8e-6 for the pre-trained model part (e.g. RoBERTa) and 5e-4 for the rest (e.g. the decoder), which was halved every 20 epochs. The weight decay during training was set to 0.01, and a dropout rate of 0.25 was applied to the decoder to prevent overfitting. The training set will be augmented during epoch 10 and epoch 20, i.e., lines 8-18 in Alg. \ref{alg:algorithm} will only be executed at epoch 10 and 20. $m$ and $n$ in Section \ref{notation} are set to 20 and 4, respectively, with the aim of pushing the limit of small math word problem solver accuracies and achieving higher performance. This is orthogonal to the contribution of this paper because our proposed customized generation strategy and exercise book are demonstrated to be effective in Sections \ref{strategy} and \ref{book}. Although we understand that further increasing the occurrences of validation and augmentation will further boost our accuracy, we limit their magnitude for better efficiency and lower cost on API calls. During validation, we generate 2 zero-shot and 2 few-shot similar problems for each source problem in the exercise book, therefore $k=4$ in Alg. \ref{alg:algorithm}. 

The backbone of our pre-trained encoder is a RoBERTa model \cite{liu2019roberta}. For LLM, we use the ChatGPT \textit{gpt-3.5-turbo} API to perform problem generation. To encourage a more diverse generation, we set the temperature to 1.25. All the experiments in this paper can be conducted with a cost lower than 100 dollars on OpenAI API calls. For evaluation, we use the value accuracy as our evaluation metric following all the baselines. 

\subsection{Comparison with Baselines}

In order to demonstrate the effectiveness of our knowledge distillation approach, we compare our method with several strong baselines, including the best fine-tuned baseline, the best LLM-enhanced knowledge distillation baseline, and chain-of-thought (CoT) prompting \cite{wei2022chain}, using the MAWPS, ASDiv-a, and SVAMP datasets. The results presented in Table \ref{tab:main_result} show that our approach outperforms all the baselines on the MAWPS and ASDiv-a datasets, achieving 94.7\% and 93.3\% solving accuracy, respectively. For the SVAMP dataset, our approach outperforms the best LLM-enhanced knowledge distillation baseline, achieving 85.4\% accuracy on the SVAMP (ID) dataset, which is a significant improvement over the prior best accuracy of 65.0\% achieved by fine-tuning. On the SVAMP (OOD) dataset, our approach achieves a solving accuracy of 76.4\%, which is lower than CoT-based LLMs, but much higher than the fine-tuned baselines. We also show the original performance of backbone solvers without any additional exercises. To obtain best performance in Table \ref{tab:main_result}, our solvers use about 20x more MWPs than the original training set to train. Overall, our results indicate that our knowledge distillation approach achieves superior performance with much fewer parameters than the prior state-of-the-art LLMs, and outperforms the best fine-tuning and knowledge distillation baselines on all datasets.

\input{./tbl_generation.tex}

\subsection{Analysis on Generation Strategy}
\label{strategy}
The performance of student solvers can be significantly impacted by the generation strategies employed to create the problems they are expected to solve. In this analysis, we explore the effectiveness of three different problem generation strategies: random, half, and target (using threshold values of $\lambda = 0, 0.5, 1$, respectively) on three datasets, in both in-distribution and out-of-distribution settings. Our goal is to identify the best strategy for maximizing student performance. We remove the initial augmentation (setting $m=0$) and lower the number of generations in this analysis to improve the efficiency of the experiments. Therefore, the results presented in Table \ref{table:generation} differ from our best results in \ref{tab:main_result}. Our analysis reveals that the targeted generation strategy generally outperforms the other two strategies. This suggests that our proposed targeted generation is an effective approach for identifying the weak areas of student solvers and improving them in a customized way. 
Moreover, we can clearly see that the improvement is more noticeable in the OOD scenario. A possible reason for this could be that in the ID situation, where the training and testing sets have some shared knowledge components, using random generation for the source problems in the training set also helps to enhance the performance on the testing set. On the other side, in the OOD scenario, where there's a large gap between the training and testing sets, our approach of creating tailored exercises specifically targets the weak points of the student model, leading to a more effective boost in its accuracy.


\input{./tbl_exercise.tex}

\begin{figure}
\centering 
\includegraphics[width=0.475\textwidth]{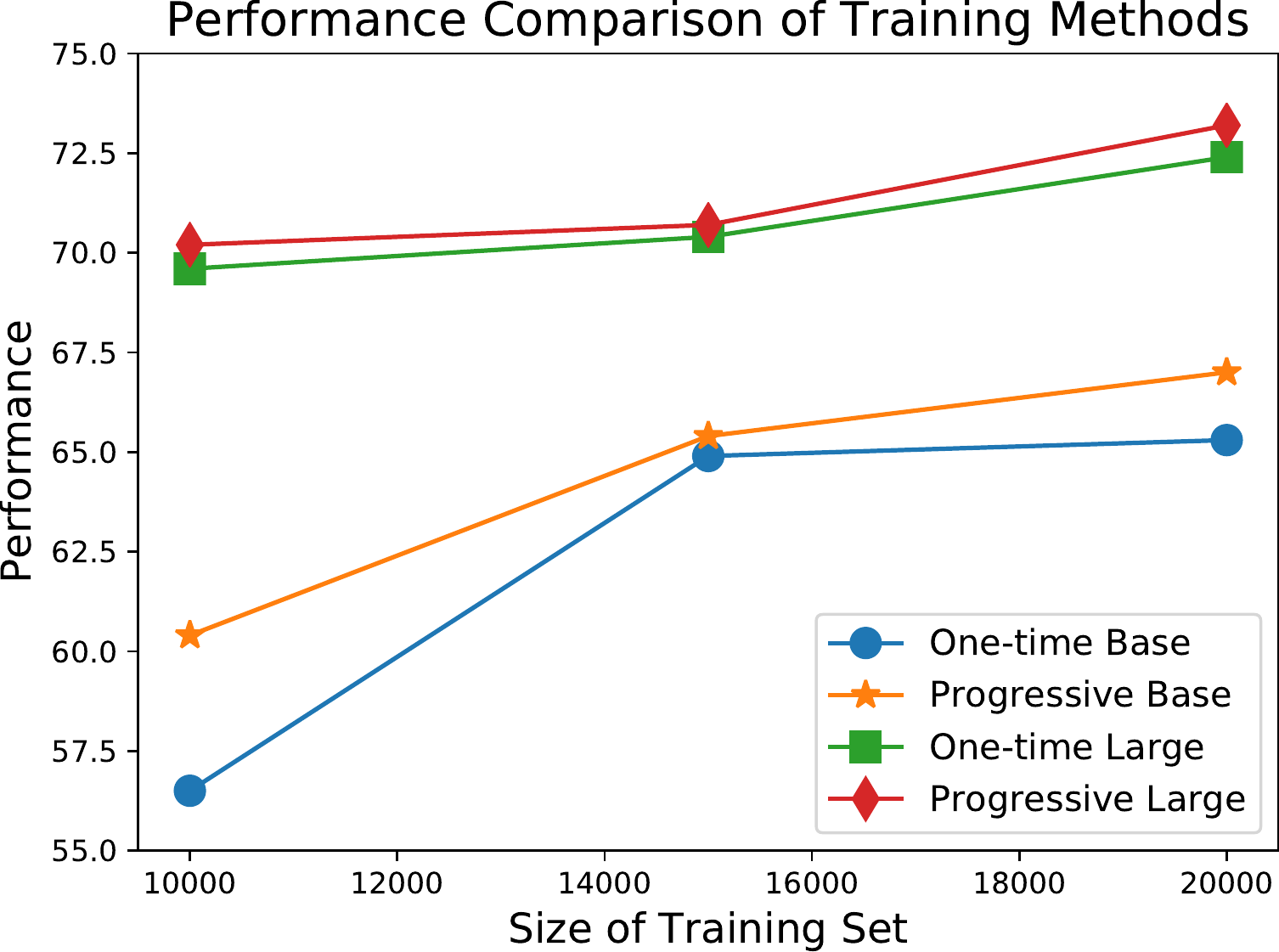} 
\caption{Performance Comparison between one-time augmentation and progressive augmentation on SVAMP under out-of-distribution setting.} 
\label{fig:aug} 
\end{figure}

\input{./tbl_case_new.tex}

\subsection{Analysis on Exercise Book}
\label{book}
We conducted an experiment to demonstrate the effectiveness of our proposed exercise book by replacing the original training set with the exercise book and changing the size $m$. As illustrated in Table \ref{tab:exercise}, the performance of generating a same-size exercise book is significantly better than that of using the training set itself for validation. Validating on the training set cannot fully reflect the learning status of the student model, which may simply memorize the solutions of training problems and is not robust to slightly perturbed problems. This strongly confirms that our proposed exercise book approach can identify weaknesses of the student model. Furthermore, by increasing the exercise book size, more knowledge can be covered, leading to further improvement in accuracy. It is worth noting that the "Same Size as Training Set" method has a similar performance to "n = 2" due to the removal of problems with wrong formats and duplicated problems, as stated in Section \ref{method}. Therefore, the size of the exercise book under these two settings is empirically similar.

\subsection{Analysis on Progressive Augmentation}
To evaluate the efficacy of our proposed progressive and customized training framework, we compare it against the one-time data augmentation approach. Specifically, one-time data augmentation means that, we augment the size of the training set at the beginning of the training process to be the same as the final size of the training set in our proposed framework and evaluate the performance of the student MWP solver on SVAMP-OOD. Intuitively, the one-time augmentation will be better than progressive augmentation because it accesses the entirely augmented training set earlier. However, as shown in Figure~\ref{fig:aug}, the results indicate that our progressive training method outperforms the one-time data augmentation approach in terms of enhancing the performance of the student solver.

\subsection{Case Study}
\label{case}
In Table \ref{tab:case}, we present several example problems generated by our method, comprising both high-quality and problematic outputs. The first case demonstrates the LLM's ability to produce a mathematically equivalent problem using distinct keywords. The second successful instance yields a sub-problem derived from the source problem, which has been proven to facilitate a better understanding of mathematical word problems for models \cite{shridhar2022automatic}. However, not all generated problems are ideal. As illustrated in the problematic example, our method occasionally generates MWPs with incorrect formatting, rendering them unsuitable for training our solver. As a result, we filter out such outputs to maintain the quality and accuracy of our training data.

\section{Conclusion}
In this work, we present a novel approach CEMAL to use large language models to facilitate knowledge distillation in math word problem solving. Our method first generates an exercise book to evaluate student models and then provides additional training exercises that are customized to the learning needs of the student model, thereby improving the student solver's ability to solve MWP. Our extensive experimental results demonstrate that our proposed approach outperforms all fine-tuned and knowledge distillation baselines on all datasets, while achieving competitive performance against LLMs with significantly fewer parameters. Additionally, we explore different selection generation strategies, revealing that our proposed customized generation strategy is the most effective method, especially in the in-distribution setting. In our future work, we aim to extend this approach to other NLP tasks to evaluate its generalization capability.

\section*{Limitations}

Despite the great performance achieved by the student model with the incorporation of our proposed technique, certain limitations remain. Firstly, our approach necessitates meticulous prompt design to generate exercises, which inevitably entails human intervention. This aspect could introduce potential bias or variability and may not scale efficiently.

Secondly, we have not explicitly addressed the quality and correctness of the generated problems. Our current filtering process only eliminates problems with incorrect formatting. Thus, there exists a significant opportunity for enhancing the effectiveness of our approach by incorporating mechanisms for evaluating and ensuring the quality and correctness of the generated exercises.

Lastly, the current framework relies on a source problem for exercise generation. Future research could explore the feasibility of generating exercises without direct reference problems, potentially utilizing only abstract knowledge components or keywords. Such an exploration could lead to better flexibility and robustness in the generation process.

\vspace{-0.45cm}

\bibliography{custom}
\bibliographystyle{acl_natbib}

\appendix



\end{document}

%% file: tbl_main.tex
\begin{table*}

\renewcommand\arraystretch{1.15}
\centering
\setlength{\tabcolsep}{1.0mm}{
\begin{tabular}{|c|c|c|c|c|c|}
\hline
\multicolumn{2}{|c|}{}           & MAWPS (ID) & ASDiv-a (ID) & SVAMP (ID) & SVAMP (OOD)    \\
\hline
 \multicolumn{2}{|c|}{Prior best (Fine-tuning)}   & $92.0/121M^a$  & $82.2/144M^b$ & $65.0/144M^b$ & $47.3/121M^a$              \\
 \multicolumn{2}{|c|}{Prior best (Knowledge Distillation)}   & $\mathit{94.5}/11.3B^c$ & $-$ & $-$ & $20.7/6.7B^d$           \\
 \multicolumn{2}{|c|}{Chain-of-Thought \cite{wei2022chain}}  & $93.3/540B$ & $\mathit{93.1}/175B^*$ & $79.0/540B$ & $\mathbf{79.0}/540B$         \\
\hline
\multirow{3}{*}{\centering \stacklines[0.2ex]{Without}{CEMAL}} & LSTM (20M) & $82.6$ & $71.4$ & $45.0$ & $30.8$ \\
& Base (144M) & $88.5$ & $81.2$ & $69.2$ & $41.0$ \\
& Large (377M) & $90.4$ & $87.6$ & $78.5$ & $49.5$ \\
\hline
\multirow{3}{*}{\centering \stacklines[0.2ex]{CEMAL}{\textbf{(Our Solvers)}}  } & LSTM (20M) & ${92.0}$ & $86.9$ & $67.1$ & $53.4$ \\
& Base (144M) & ${93.9}$ & $90.9$ & $\mathit{81.5}$ & $68.6$ \\
& Large (377M) & $\mathbf{94.7}$ & $\mathbf{93.3}$ & $\mathbf{85.4}$ & $\mathit{76.4}$ \\
\hline
\end{tabular}}
\caption{We compare the accuracy and number of parameters on 4 benchmarks in the format of (accuracy/number of parameters). Prior best baselines are the following. a: \cite{jie2022learning}, b: \cite{patel2021nlp}, c: \cite{magister2022teaching}, d: \cite{ho2022large}. On each dataset, the best performance is \textbf{bolded} and the second best is in $\mathit{italics}$. $^*$: This accuracy is calculated on the ASDiv-a subset out of ASDiv based on the results in \url{https://github.com/jasonwei20/chain-of-thought-prompting}. ID denotes in-distribution test and OOD denotes out of distribution test.}
\label{tab:main_result}
\end{table*}

%% file: tbl_generation.tex
\begin{table}[t]
\renewcommand\arraystretch{1.05}
\centering
\setlength{\tabcolsep}{0.85mm}
\resizebox{\columnwidth}{!}{
\begin{tabular}{ccccc}
\hline
\multirow{2}{*}{Dataset} & \multirow{2}{*}{Backbone} & \multicolumn{3}{c}{In-Distribution} \\
& & Random & Half & Target \\
\hline
\multirow{3}{*}{\centering ASDiv-a} & GTS & 77.5 & 77.8 & \textbf{79.0} \\
& RoBERTa-base & 84.3 & \textbf{85.2} & 84.6 \\
& RoBERTa-large & 90.1 & 90.3 & \textbf{90.6} \\
\hline
\multirow{3}{*}{\centering MAWPS} & GTS & 89.2 & 88.9 & \textbf{89.3} \\
& RoBERTa-base & 90.3 & 90.3 & \textbf{91.0} \\
& RoBERTa-large & 91.4 & 92.3 & \textbf{92.8} \\
\hline
\multirow{3}{*}{\centering SVAMP} & GTS & 47.7 & 49.6 & \textbf{50.3} \\
& RoBERTa-base & 72.1 & 72.8 & \textbf{73.3} \\
& RoBERTa-large & 80.8 & 79.9 & \textbf{80.9} \\
\hline

\multirow{2}{*}{Dataset} & \multirow{2}{*}{Backbone} & \multicolumn{3}{c}{Out-of-Distribution} \\
& & Random & Half & Target \\
\hline
\multirow{3}{*}{\centering SVAMP} & GTS & 33.6 & 36.3 & \textbf{38.2} \\
& RoBERTa-base & 50.1 & 50.8 & \textbf{55.3} \\
& RoBERTa-large & 62.9 & 63.1 & \textbf{65.0} \\
\hline
\end{tabular}}
\caption{Results of different problem generation strategies on three datasets under in-distribution and out-of-distribution (OOD) testing. Boldface indicates the best result among the three strategies.}
\label{table:generation}
\end{table}

%% file: tbl_exercise.tex
\begin{table}
\centering
\resizebox{\columnwidth}{!}{%
\begin{tabular}{ccccc}
\toprule
& \makecell{Training set \\ as exercise book} & \makecell{Same size \\ as training set} & \makecell{n = 2} & \makecell{n = 4} \\
\midrule
Base & 54.6 & 59.8 & 60.4 & 65.2 \\
Large & 64.7 & 69.8 & 70.2 & 72.7 \\
\bottomrule
\end{tabular}%
}
\caption{Performance comparison of using training set and different sizes of exercise book for validation on SVAMP dataset.}
\label{tab:exercise}
\end{table}

%% file: tbl_case_new.tex
\begin{table*}[t]
\renewcommand\arraystretch{1.15}
\centering
\begin{tabular}{|p{1.6cm}|p{13.5cm}|}
\hline
\multicolumn{2}{|>{\columncolor{good_color}}c|}{\textbf{Case 1: Equivalent problem generation}} \\
\hline
\multicolumn{2}{|c|}{\cellcolor{good_color}This case shows that LLM generates an equivalent problem using different keywords.} \\
\hline
\textbf{Source Problem} & In a school there are N0 girls and N1 boys. How many more girls than boys does the school have? \\
\hline
\textbf{Answer} & N0 - N1 \\
\hline
\textbf{Generated Problem} & In a zoo there are N0 lions and N1 tigers. How many more lions than tigers are there in the zoo? \\
\hline
\textbf{Answer} & N0 - N1 \\
\hline
\multicolumn{2}{|>{\columncolor{good_color}}c|}{\textbf{Case 2: Sub-problem generation}} \\
\hline
\multicolumn{2}{|c|}{\cellcolor{good_color}This case demonstrates an instance where a sub-problem is derived from the source problem.} \\
\hline
\textbf{Source Problem} & In a school there are N0 girls and N1 boys. N2 more girls joined the school. How many more girls than boys does the school have now? \\
\hline
\textbf{Answer} & N0 + N2 - N1 \\
\hline
\textbf{Generated Problem} & In a school there are N0 girls and N1 boys. N2 more girls joined the school. How many girls are there in total now? \\
\hline
\textbf{Answer} & N0 + N2 \\
\hline
\multicolumn{2}{|>{\columncolor{bad_color}}c|}{\textbf{Case 3: Incorrect format generation}} \\
\hline
\multicolumn{2}{|c|}{\cellcolor{bad_color}This case shows a problem where the LLM generates a problem with incorrect formatting.} \\
\hline
\textbf{Source Problem} & For Gwen's birthday she received N0 dollars. She spent some money and has N1 dollars left. How much money did she spend? \\
\hline
\textbf{Answer} & N0 - N1 \\
\hline
\textbf{Generated Problem} & For Gwen's birthday she received N0 dollars. She spent some money and now has N1 dollars left. How much money did she originally have? \\
\hline
\textbf{Answer} & N0 = N1 + spent amount (Wrong format) \\
\hline
\end{tabular}
\caption{Examples of problem generation by our method. Case 1 and Case 2 (colored in green) show successful instances where mathematically equivalent problem and sub-problem are generated respectively. Case 3 (colored in red) illustrates a problem with incorrect formatting.}
\label{tab:case}
\end{table*}